# Design and Experimental Evaluation of a Haptic Robot-Assisted System for Femur Fracture Surgery*

Fayez H. Alruwaili, Michael P. Clancy, Marzieh S. Saeedi-Hosseiny, Jacob A. Logar, Charalampos Papachristou, Christopher Haydel, Javad Parvizi, Iulian I. Iordachita, Mohammad H. Abedin-Nasab

***Abstract* — In the face of challenges encountered during femur fracture surgery, such as the high rates of malalignment and X-ray exposure to operating personnel, robot-assisted surgery has emerged as an alternative to conventional state-of-the-art surgical methods. This paper introduces the development of Robossis, a haptic system for robot-assisted femur fracture surgery. Robossis comprises a 7-DOF haptic controller and a 6-DOF surgical robot. A unilateral control architecture is developed to address the kinematic mismatch and the motion transfer between the haptic controller and the Robossis surgical robot. A real-time motion control pipeline is designed to address the motion transfer and evaluated through experimental testing. The analysis illustrates that the Robossis surgical robot can adhere to the desired trajectory from the haptic controller with an average translational error of 0.32 mm and a rotational error of 0.07°. Additionally, a haptic rendering pipeline is developed to resolve the kinematic mismatch by constraining the haptic controller's (user's hand) movement within the permissible joint limits of the Robossis surgical robot. Lastly, in a cadaveric lab test, the Robossis system assisted surgeons during a mock femur fracture surgery. The result shows that Robossis can provide an intuitive solution for surgeons to perform femur fracture surgery.**

## I. INTRODUCTION

Robot-assisted surgery has surged in the spotlight of surgical technology due to the demand for increased accuracy and speed needed for the evolution of healthcare [1]–[4]. This evolution has impacted many targetable areas in need of improvement for ideal surgical outcomes [1]–[4]. Blood loss, incision size, surgical time, and hospital stay duration can all be minimized to improve patient outcome. However, surgical robotics tends to lack the surgeon's connection to the operation, creating a gap in the intuitive skill set required for successful operations. The gap is seen through the lack of the sense of touch, which can lead to tissue damage due to the inability to identify tissue consistency and inconsistency of applying forces through the robot [5]–[10]. Also, the lack of sensory feedback creates challenges in invasive surgeries, such as the inability to make accurate intricate surgical tasks/movements, prevent surgical tremors, or determine the shape/boundary of the surgical workspace [11], [12].

The introduction of haptic feedback technology in surgical robotics has demonstrated the ability to eliminate the gap in intuitive skills necessary for improved surgical outcomes. Haptics have proved to enhance surgeon-robot interactions by providing somatic sensory feedback along with visual cues [9]–[11], [13]. Furthermore, haptic feedback has improved the safety and efficacy of surgical robotics by defining a safe surgical workspace, improving user maneuverability, decreasing user response time, and modeling tissue consistency and interaction through applied torques and forces [9], [10], [13].

Haptics and surgical robots were first implemented in a prototype surgery in 2015 when Senhance's Telelap Alf-x used force sensors on the end of laparoscopic instruments to transmit grasping force during a cholecystectomy [14]–[17]. Since then, interest in the implementation of haptic feedback into surgical robotics has grown drastically in many surgical specialties [18]–[20]. For example, some groups have incorporated haptic feedback with the traditional da Vinci robot [21], [22].

Challenges faced during femur fracture surgery have positioned robot-assisted surgery as a potential alternative to the traditional state of the art clinical techniques [23], [24]. These challenges include the substantial amount of force required to reposition and fixate the bone fragments, unacceptably high rates of malalignment and malrotation, and the high X-ray exposure to the operating staff [23], [24]. Numerous research studies have been conducted on developing a robot-assisted system for femur fracture surgery, but no system has yet been adequately designed to meet the clinical and mechanical requirements specific to femur fracture surgery applications [23]–[28]. Particularly, the need for immense force competence, 517 N and 74 N·m [29], [30], pinpoint accuracy, Thoresen scoring system as: $\pm 1cm$ translational and $\pm 5°$ rotational alignment [31], the required surgical-robot workspace, and the benefit of haptic feedback.

*This work is funded by the National Science Foundation (NSF) under grants 2141099, and 2226489, and by the New Jersey Health Foundation (NJHF) under grant PC 62-21. *(Corresponding author: Mohammad H. Abedin-Nasab)*
F. Alruwaili, M. Clancy, J. Logar, and M. H. Abedin-Nasab are with the Biomedical Engineering Department, Rowan University, Glassboro, NJ 08028, USA (e-mail: alruwa16@rowan.edu, clancym1@rowan.edu, logarj15@rowan.edu, corresponding author email: abedin@rowan.edu).
M. S. Saeedi-Hosseiny is with the Electrical and Computer Engineering Department, Rowan University, Glassboro, NJ 08028, USA (e-mail: saeedi64@rowan.edu).
C. Papachristou is with the College of Science & Mathematics, Rowan University, Glassboro, NJ 08028 (e-mail: papachristou@rowan.edu).
C. Haydel is an Orthopedic Trauma Surgery with Virtua Health, Moorestown, NJ 08057 (e-mail: chaydel@virtua.org).
J. Parvizi is with Rothman Orthopedic Institute, Thomas Jefferson University Hospital, Philadelphia, Pennsylvania (e-mail: javadparvizi@gmail.com).
I. I. Iordachita is with the Laboratory for Computational Sensing and Robotics, Johns Hopkins University, Baltimore, MD 21218 USA (e-mail: iordachita@jhu.edu).

Chan et al. [32] and Kong et al. [33] developed teleoperation robot systems for femur fracture alignment based on the Gough-Stewart platform (GSP). Their systems include a bilateral force feedback control framework, a tracking system, and bone-fixing support. However, the main challenge that hinders the developed systems is their practical use in the clinical setting. The small translational and rotational workspaces of the GSP, in combination with the large mechanism size, limit the design's efficacy for femur fracture applications [34]. Furthermore, Ralf et al. [35] developed a teleoperation robot system based on the serial-robot mechanism and 6 DOF haptic joystick. A limitation of the robot's serial connection structure is that it is unable to produce the force needed to withstand the large muscle loads around the femur while maintaining an appropriate size for use in the operating room [23], [24].

To address these challenges faced during long-bone fracture surgeries, Abedin-Nasab et al. [36] presented a novel 3-armed 6-DOF parallel robot, called the Robossis Surgical Robot (RSR). The previous theoretical and experimental analysis demonstrates that the RSR has a rotational workspace 15 times larger than the GSP and can generate the required traction forces up to 1,100 N with sub-millimeter accuracy [37]–[42]. Furthermore, preclinical cadaver testing demonstrates the feasibility of the RSR to be used in a clinical setting [43] [44].

In this paper, we integrate a Haptic Controller (HC), named the leader, to the RSR, named the follower, to (i) provide an intuitive method to manipulate the robot and (ii) apply force feedback to the surgeon's hand. To this end, we develop a unilateral control architecture that includes (i) a control unit to transfer the motion from the HC to the RSR and (ii) a haptic rendering pipeline that addresses the kinematic mismatch between the HC and RSR. The key aspects of our development in this paper are the following:

1. Propose a unilateral control architecture that integrates the HC with the RSR. The unilateral control architecture addresses the challenges of integrating a leader-follower system for kinematically dissimilar robots and motion transfer while maintaining a safe and minimal delay.
2. Develop a real-time motion control pipeline for the motion transfer from the HC to RSR. We present the integration of the Kalman filter to predict the state of the RSR while maintaining a continuous position, velocity, and acceleration profile. Also, we propose a robust dynamic scaling factor that limits the instantaneous velocity of the RSR trajectory to meet the required safety limits.
3. Develop a haptic rendering pipeline to resolve the kinematic mismatch by constraining the HC's (user's hand) movement within the permissible joint limits of the RSR. We model the HC end effector by two spherical point clouds to determine the interaction with the workspace surfaces. The users experience haptic feedback in the directions of the workspace edges to keep the HC inside the RSR workspace limits.

Furthermore, we design a virtual robot within the Gazebo environment to validate the kinematics of the mechanism and the relationship between users' desired motion and input trajectory. Also, we evaluate the performance of the proposed work via theoretical and experimental testing. Lastly, we conduct a cadaver experiment to demonstrate the feasibility of the Robossis system.

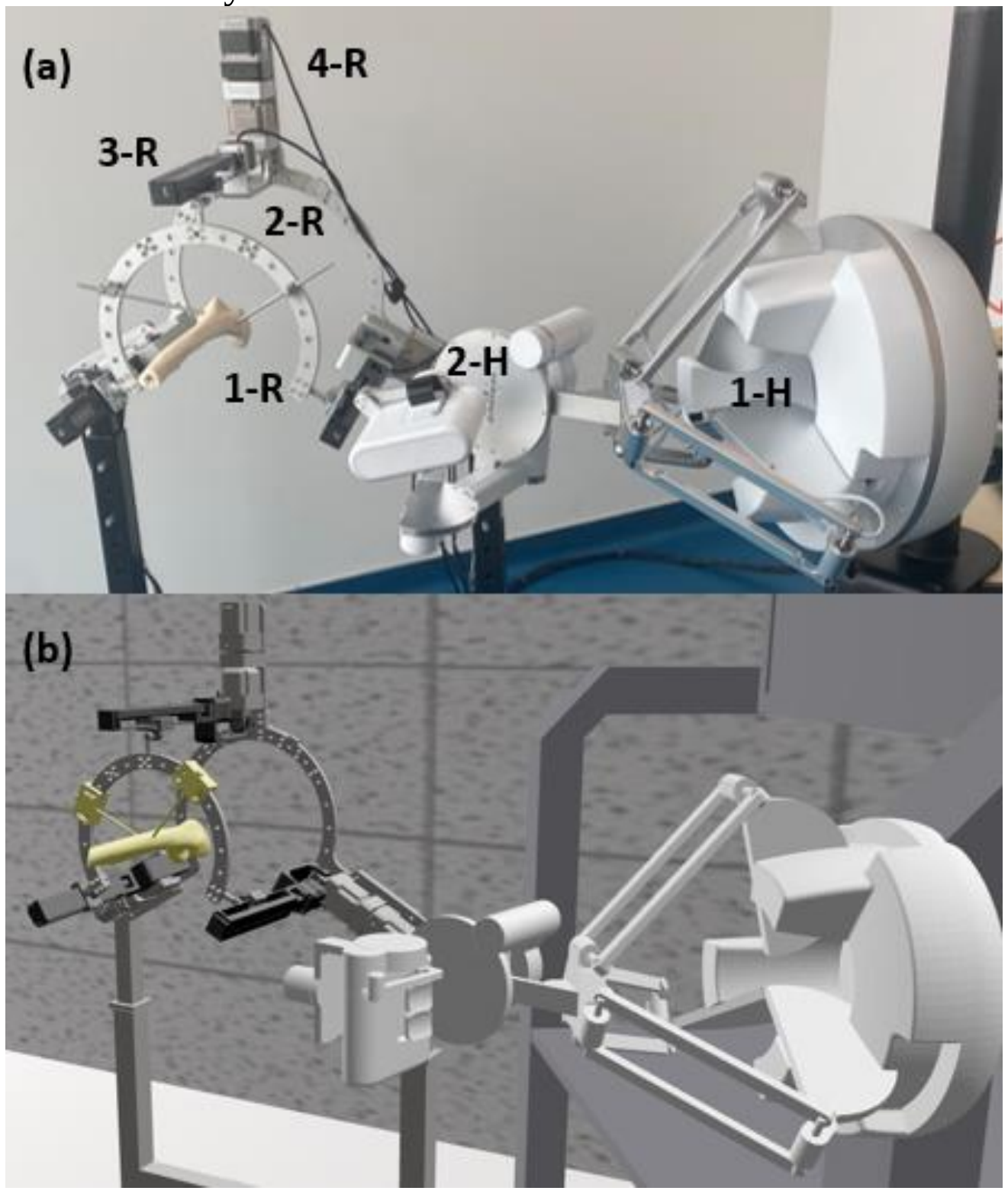

**Fig. 1. (a)** Real-life representation of the Robossis system. The HC consists of three identical legs connected in parallel (1-H), attaching between a top triangular fixed base and a mobile end-effector platform and coupled with a hybrid serial arm (2-H). The RSR consists of a moving ring (1-R), a fixed ring (2-R), and three arms, where each arm consists of a linear (3-R) and rotary actuator (4-R). **(b)** Identical representation of the Robossis system in the Gazebo simulator.

The remainder of this paper is organized as follows. The architecture of the Robossis system is described in Section II. Section III provides the kinematics of the HC and RSR. In Section IV, the virtual robot simulation and analysis are presented. The motion transfer between the HC and RSR is proposed in Section V. The haptic rendering pipeline is proposed in Section VI. Experimental and cadaver testing to validate the performance of the HC and RSR is presented in Section VII. Finally, the discussion and conclusion are presented in Sections VIII and IX.

## II. Robossis System Architecture

The architecture of the Robossis system is described in this section. The Robossis system includes a 7-DOF HC (Sigma.7, Force Dimension - Switzerland) and the 6-DOF RSR. The sigma.7 HC is a hybrid robot structure based on a delta mechanism providing 3-DOF translational manipulation, a wrist serial mechanism providing 3-DOF rotational manipulation, and a grasping unit for 1-DOF [45], [46] (**Fig. 1).**

The RSR is designed based on a 3-armed parallel mechanism where each arm is placed on a moving and fixed ring (**Fig. 1**). Each arm is actuated with a linear and rotatory actuator (**Fig. 1**). The Robossis system is designed to meet the clinical requirements of femur fracture surgery, which includes 1) adequately applying the large traction forces/torques, 2) precisely aligning the fractured bone, and 3) manipulating the distal bone fragment during the surgical procedure (in our previous work we provide a clear description of the mechanical and clinical requirement of the RSR [39], [43]). Also, **Fig. 1 (b)** shows the identical representation of the Robossis system in the Gazebo simulator.

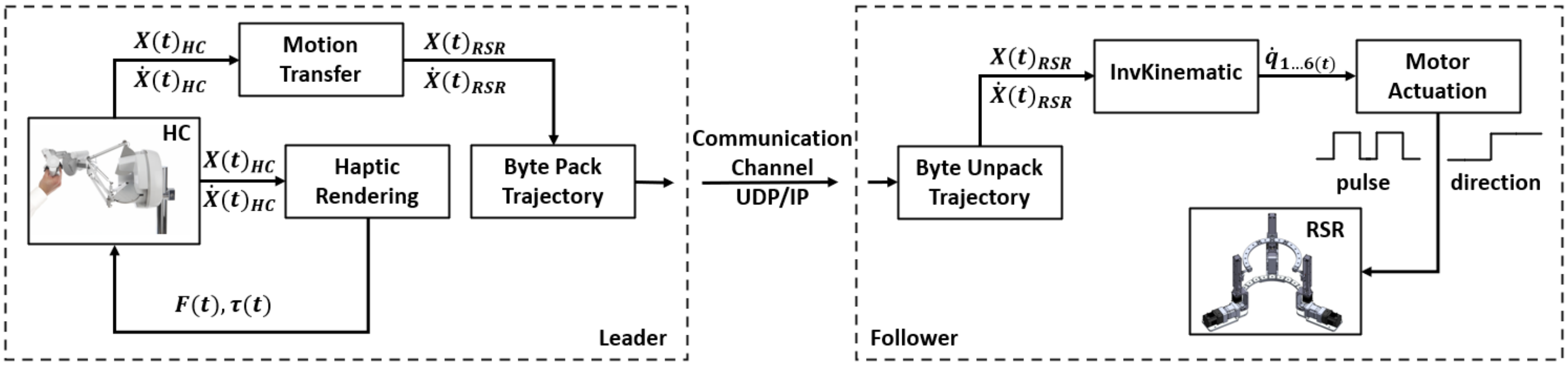

**Fig. 2.** The proposed Robossis system unilateral architecture is illustrated. A leader unit is established where the haptic rendering loop is included to ensure safety for the manipulation of the RSR within its joint space limits. Also, the motion transfer is included that integrates the motion of the user's hand to the RSR for a continuous position, velocity, and acceleration trajectories. A communication channel is established between the leader and follower. In the follower unit, the RSR joint velocities are estimated and the low-level communication with the RSR actuators as pulses and direction are established.

The entire architecture includes a leader unit, a follower unit, and a communication channel (**Fig 2**). First, in the leader unit, the user interacts with the HC, and the instantaneous position and velocity are measured from the HC as $\boldsymbol{x(t)_{HC}} \in R^6$ and $\boldsymbol{\dot{x}(t)_{HC}} \in R^6$, respectively. The architecture implements a force restriction, $\boldsymbol{F(t)} \in R^3$ & $\boldsymbol{\tau(t)} \in R^3$**,** on the HC movement to be restricted inside the changing allowable workspaces of the RSR. These restrictions ensure that the movement of the HC is within the boundaries of the RSR and eliminate any damage to the RSR actuators and structure. Second, motion transfer between the HC and RSR is implemented to map the position and velocity of the users' (surgeons') to the end effector of the RSR as $\boldsymbol{x(t)_{RSR}} \in R^6$ and $\boldsymbol{\dot{x}(t)_{RSR}} \in R^6$, respectively. We implement the Kalman filter to predict the state of the RSR with continuous position, velocity, and acceleration trajectories. Additionally, we implement a robust dynamic scaling method that restricts the input trajectories to the maximum linear and angular velocity (user-defined). Furthermore, we establish a communication channel between the leader and follower units. In the follower unit, the trajectories are unpacked, and the RSR joint velocities are estimated as $\boldsymbol{\dot{q}(t)} \in R^6$. We handle the low-level communication between the control software and the RSR actuators as pulses and directions.

## III. Inverse Kinematic Analysis

### *A. Inverse Kinematics: Haptic Controller*

Based on the previous kinematic analysis of the Falcon 3-DOF haptic controller [47], [48], a kinematic analysis for the sigma.7 is presented here. **Fig. 3** presents a top-side view schematic for the three kinematic chains of the sigma.7 HC and its dimension, where i denotes the ith kinematic chain. The origin (center of the stationary platform) is attached to the coordinate frame (x, y, z) and labeled point $\boldsymbol{O}$, and point $\boldsymbol{P}$ is defined as the center of the moving platform. The links' lengths are labeled "a" (84 mm) and "b" (175 mm), while the distance between the lowest joint $A_i$ and the center of the stationary platform is indicated by "r" (79 mm). Furthermore, "c" (42 mm) denotes the distance of the highest joint $E_i$ to the center of the moving platform, and "f" (37 mm) denotes the length to the center of the moving platform. A coordinate frame $(u_i, v_i, w_i)$ is defined for each kinematic chain, attached at point $A_i$ (**Fig. 3**).

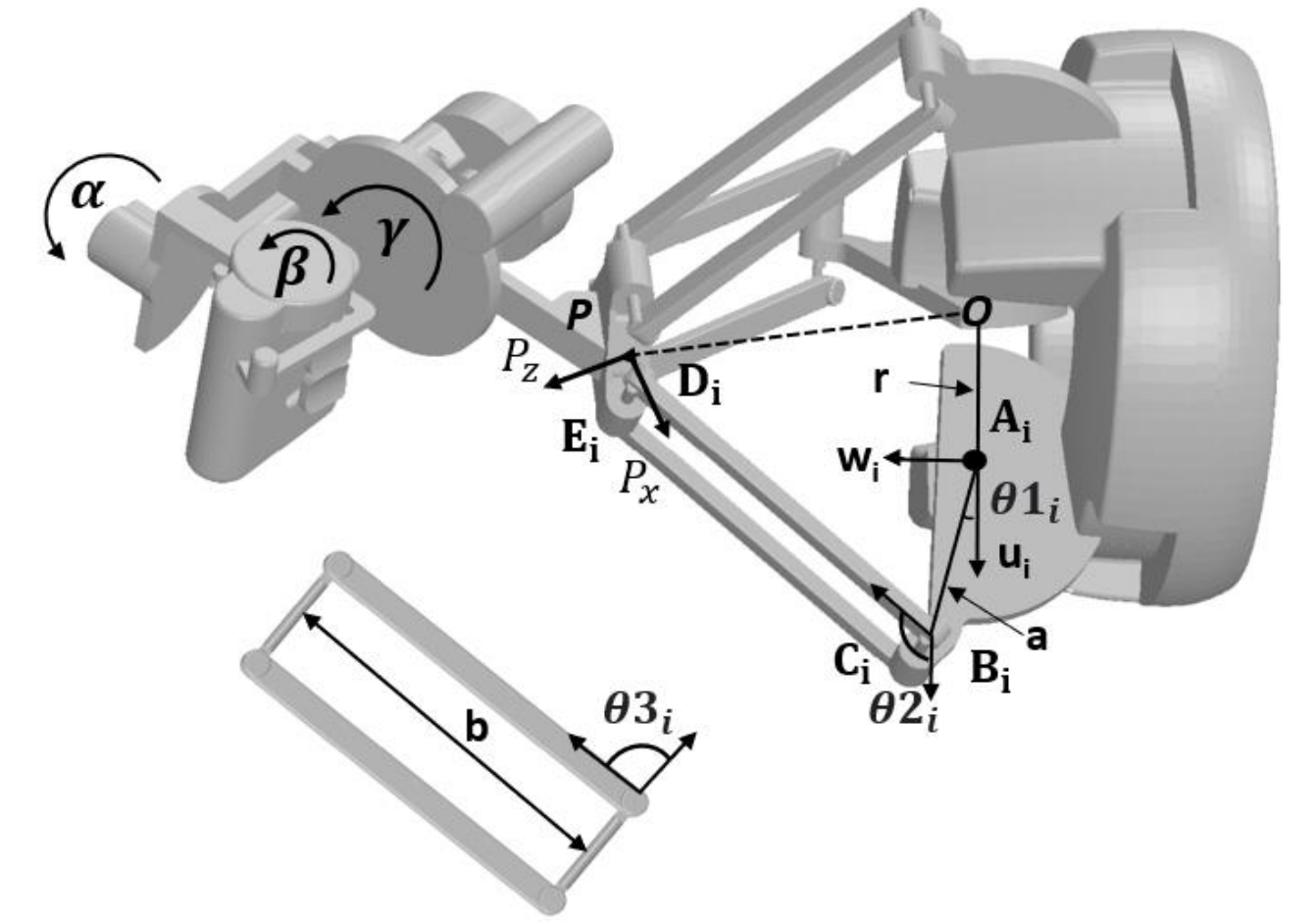

**Fig. 3.** Kinematic variables of the HC delta mechanism ith arm are shown. The serial mechanism has 3 independent axes of rotation.

Equation (1) presents the coordinate transformation of the position of point **P** in the $(u_i, v_i, w_i)$ coordinate system:

$$\begin{bmatrix} Pu_i \\ Pv_i \\ Pw_i \end{bmatrix} = \begin{bmatrix} \cos(\varphi_i) & \sin(\varphi_i) & 0 \\ -\sin(\varphi_i) & \cos(\varphi_i) & 0 \\ 0 & 0 & 1 \end{bmatrix} \begin{bmatrix} \cos(\theta) & 0 & \sin(\theta) \\ 0 & 1 & 0 \\ -\sin(\theta) & 0 & \cos(\theta) \end{bmatrix} \begin{bmatrix} Px \\ Py \\ Pz \end{bmatrix} + \begin{bmatrix} -r \\ -s \\ 0 \end{bmatrix} \quad (1)$$

where $\theta$ is 45°, and the angle of the $u_i$-axis, $\varphi_i$, is the distance of the rotation joint from the y-axis. $\varphi_i$ evaluates as $-17°, -137°,$ and $103°$. From **Fig. 3**, an expression for $(Pu_i, Pv_i, Pw_i)$ is derived as below (2)-(4):

$$Pu_i = a\cos(\theta_{1i}) - c + [b\sin(\theta_{3i})]\cos(\theta_{2i}) \quad (2)$$

$$Pv_i = b\cos(\theta_{3i}) - f \quad (3)$$

$$Pw_i = a\sin(\theta_{1i}) + [b\sin(\theta_{3i})]\sin(\theta_{2i}) \quad (4)$$

For each x, y, and z position, four solutions could be obtained. Due to the physical constraints of the manipulator, joint $\theta_{3i}$ can only have positive values, which consequently eliminates two solutions. Further, $\theta_{1i}$ is always in the first coordinate, eliminating the remaining solutions, and this leaves one real solution.

### *B. Inverse Kinematics: Robossis Surgical Robot*

We map the input of the user's hand location and orientation (**Fig. 4**), $\boldsymbol{P} \in R^6$, as the desired location of the RSR end effector (center of the moving ring ($\boldsymbol{P}$)). Given the position ($\mathrm{P}(\mathrm{x}, \mathrm{y}, \mathrm{z})$) and orientation ($\mathrm{R}(\alpha, \beta, \gamma)$) of the endpoint effector ($\boldsymbol{P}$), the length of the linear actuator ($d_i$) and the rotation of the active joint ($\theta_i$) are computed. Referring to **Fig. 4**, $\mathbf{a_i}$ and $\mathbf{b_i}$ represent $OA_i$ and $PB_i$, respectively.

Denoting the $\mathbf{a_i}$ and $\mathbf{b_i}$ position vector in frame {A}, it can be concluded from the structure that

$$\boldsymbol{r_i} - \boldsymbol{a_i} = \boldsymbol{p} + \boldsymbol{b_i} - \boldsymbol{a_i} \quad (5)$$

where the left-hand side is the length vector of the linear actuator ($\boldsymbol{d_i}$). Simplifying and using Euclidean norms, $d_i$ can be expressed as

$$d_i = \sqrt{(x - x_i)^2 + (y - y_i)^2 + (z - z_i)^2} \quad (6)$$

The active ($\theta_i$) and passive ($\psi_i$) is expressed in (3) and (4) as:

$$\theta_i = \sin^{-1}\left(\frac{\sin(\gamma_i) * (x - x_i) - \cos(\gamma_i) * (y - y_i)}{d_i cos * (\psi_i)}\right) \quad (7)$$

$$\psi_i = \sin^{-1}\left(\frac{\cos(\gamma_i) * (x - x_i) + \sin(\gamma_i) * (y - y_i))}{d_i}\right) \quad (8)$$

where $\gamma_i$ is the location of each of the ith arms on the moving platform and denoted as -30°, 90°, and 210°. Physical constraints, including the length of linear actuators (± 70 mm), and spherical joints (± 25°), are imposed to determine the allowable translational and rotational workspace of the RSR structure.

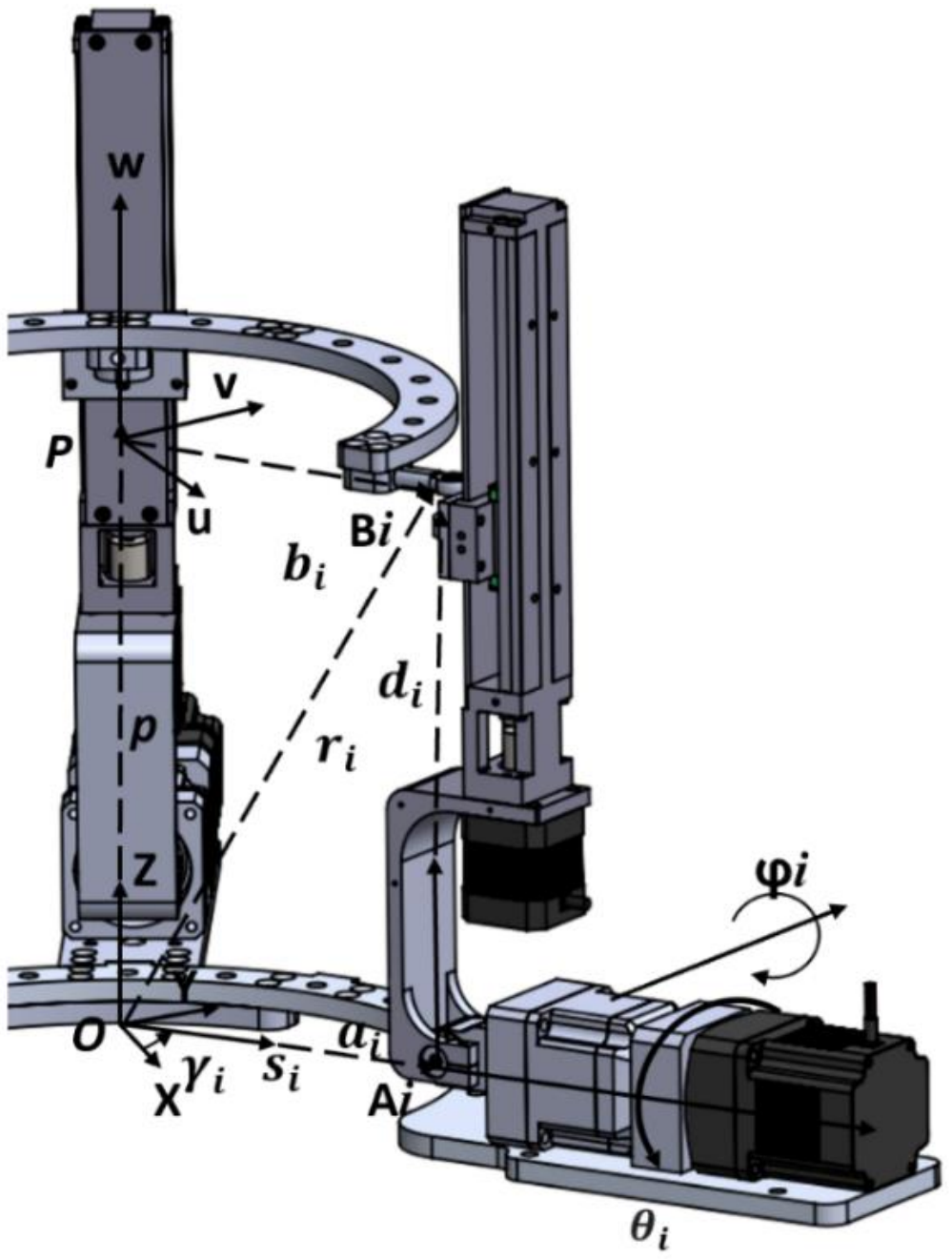

**Fig. 4.** Kinematic variables of the ith arm of the RSR are shown. $d_i$ is the linear actuator length, $\theta_i$ is the active rotation, followed by the passive $\psi_i$ rotation.

### *C. Kinematic Analysis: Joint Space Limits*

The dexterous and rotational workspaces for the RSR and HC were constructed given the joints' dimensions and constraints (**Fig. 5).** The HC's dexterous workspace presents a cone-like shape and it's independent of the position and orientation of the end-effector. Whereas the RSR dexterous workspace presents a dome-like shape and it's dependent on the position and orientation of the mechanism end-effector. Furthermore, the theoretical rotational workspace of RSR presents a diamond-like structure, whereas sigma.7 is based on a serial arm, so it is a grid-like structure with the following limits (235° x 140° x 200°). The kinematic analysis illustrates the kinematic dissimilarities between the HC controller and RSR where each structure presents a unique operational dexterous and rotational workspace. Also, the RSR operational dexterous and rotational workspaces change as a function of end effector position and orientation whereas HC workspaces remains unchanged due to the hybrid nature of the HC mechanism.

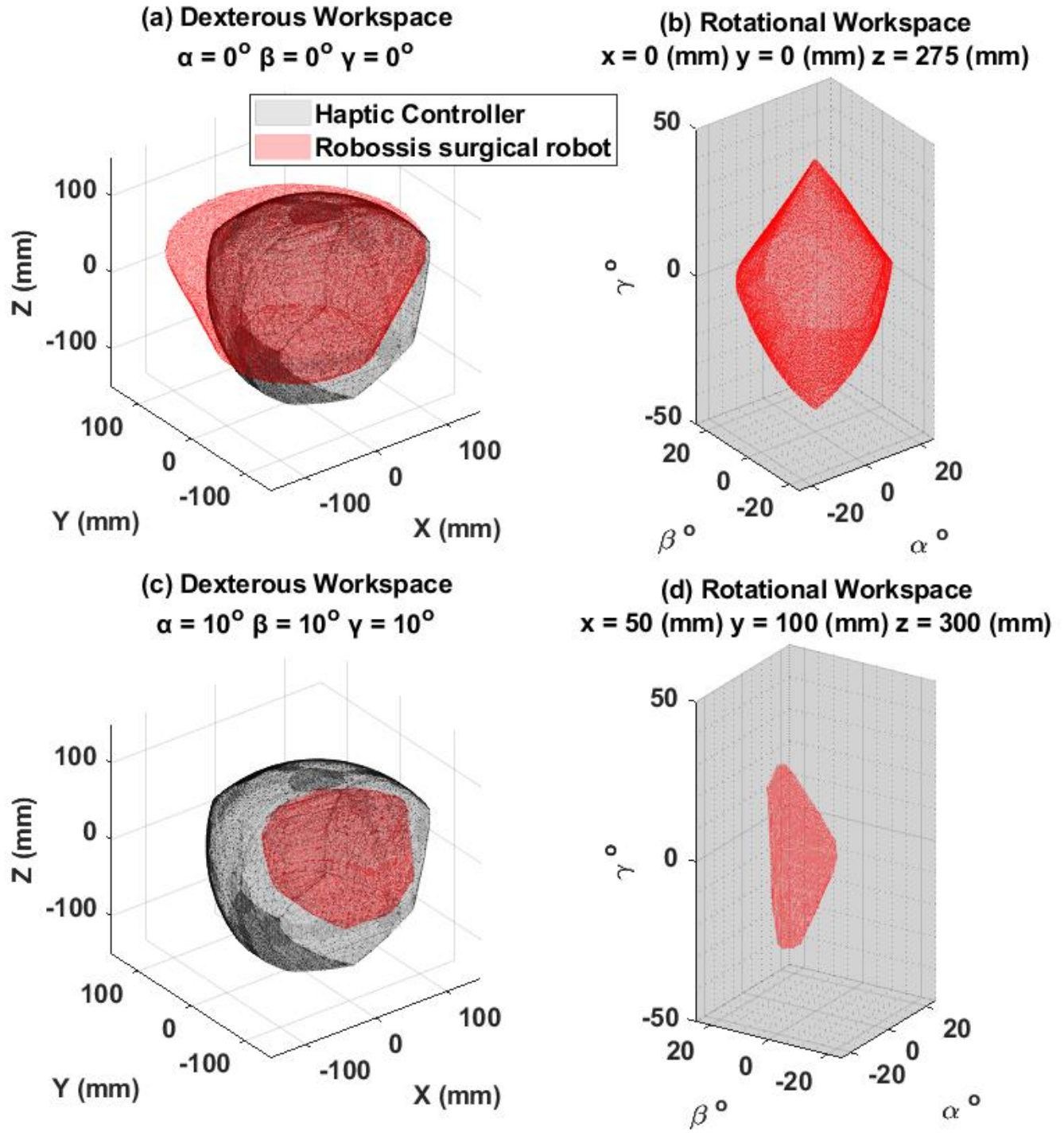

**Fig. 5.** Theoretical simulation of the HC and RSR operational workspace for the dexterous and rotational workspace. **(a)** shows the dexterous workspace of RSR and HC at constant orientation of the end effector. **(b)** demonstrate the rotational workspace of RSR and HC at constant position of the end effector. **(c)** and **(d)** present the dexterous and rotational workspace at a different position and orientation of the end effector. The RSR operational dexterous and rotational workspaces change as a function of the end effector position and orientation whereas the HC remains unchanged due to the hybrid nature of the HC mechanism. The rotational workspace for sigma.7 is based on a serial arm, so it is a grid-like structure with the following limits (235° x 140° x 200°).

## IV. Virtual Robot Simulation and Analysis

To verify the kinematics of the RSR mechanism and input-output relationship, we develop a virtual RSR using the Gazebo simulator. The robot was represented in the simulator by dividing its components into different links (L1-L5) and connecting them through joints (J.Ai-Di), following the parent-child convention to establish link relationships (as shown in **Fig. 6**). Each arm of the robot consisted of three joints: active and passive universal joints (J.Ai & J.Bi), prismatic joint (J.Ci), and spherical joint (J.Di) (as depicted in **Fig. 6**). The universal joint (L2i) connected the rotary actuator shaft (L1) to the lower arm (L3i), which was fixed to the platform. Similarly, the spherical joint connected the upper parts of the linear actuators (L4i) to the moving ring (L5). This resulted in a closed-loop link-joint relationship in Gazebo that closely resembled the real-world configuration of the Robossis surgical robot.

A simulation inside the gazebo simulator was performed to determine the RSR end effector deviation from the input desired path. During the simulation, a sinusoidal path with varying periods for each axis was input as the desired path for RSR to follow for all 6-DOF (translation and rotation) (video attached). During the simulation, the lengths of the linear actuators and the angles of the active joints' were calculated and interfaced with the RSR in the Gazebo simulator using a custom model plugin. **Fig. 7** displays the desired and measured input path for translation (**Fig.7 a**) and rotation (**Fig.7 b**), as well as the corresponding error observed during the simulation (**Fig.7 c**). The maximum deviation observed for the translational and rotational motion was ~ 0.2 mm and ~ 0.1 degrees, respectively. As such, the simulation validates the kinematics of the RSR to accurately follow a desired trajectory and defines the relationship between the users' desired motion and input trajectory.

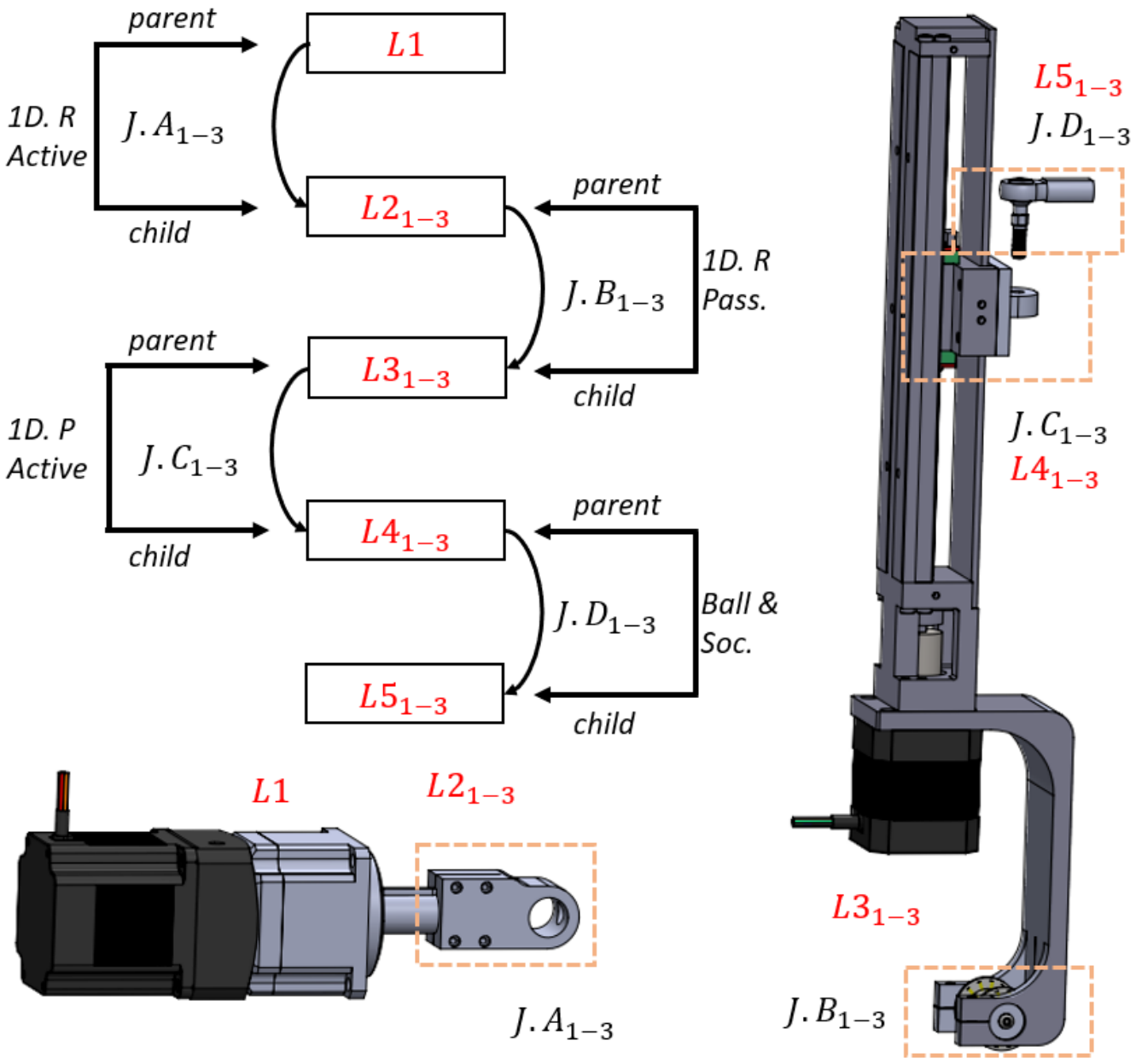

**Fig. 6.** RSR representation within the Gazebo environment as links and connected via the parent-child joint hierarchy.

## V. Motion Transfer and Control

Motion transfer and control of the RSR can be achieved with a manual graphical user interface (GUI) or real-time motion via the HC (sigma.7). For manual motion, a time-controlled trajectory generation scheme is implemented using a trigonometric function as:

$$\boldsymbol{x(t)_{RSR}} = \begin{cases} x_0, \ \ t < t_0 \\ x_0 + dx, \ \ t > t_0 + dt \\ \frac{dx}{2} * sin\left(\frac{180}{dt} * (t - t_0) - 90\right) + x_0 + \frac{dx}{2} \end{cases} \quad (9)$$

where $x_0$ is the initial position, $dx$ is the desired change of motion, $t_0$ is the initial time of change, and $dt$ is the desired time to complete the motion. The time-controlled trajectory scheme is generalized to all translational and rotational motions.

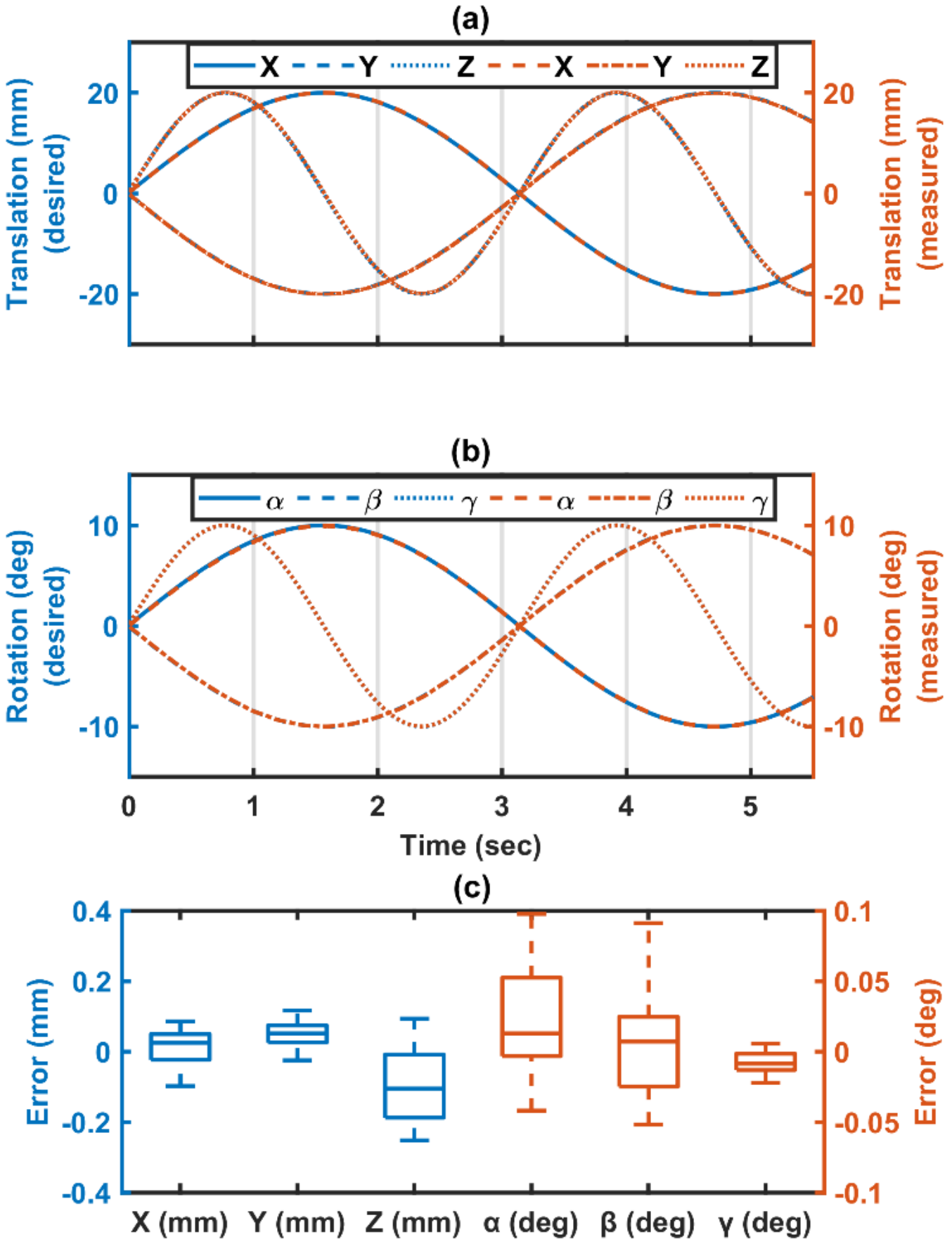

**Fig. 7.** Robossis surgical robot's desired trajectory as commanded a sinusoidal trajectory in 6 DOF (**(a)** translation and **(b)** rotation). **(c)** A maximum deviation for the translational motion (~ 0.2 mm) and rotation motion (~ 0.1 deg) during the entire simulation.

Furthermore, the real-time motion integration between the HC (user's hands) and RSR is achieved as an incremental trajectory as

$$\boldsymbol{x(t)_{RSR}} = \boldsymbol{x_{RSR}(t-1)} + (\boldsymbol{x(t)_{HC}} - \boldsymbol{x(t-1)_{HC}}) * \boldsymbol{S} \quad (10)$$

where $\boldsymbol{x(t)_{RSR}} \in R^6$ and $\boldsymbol{x_{RSR}(t-1)} \in R^6$ is the current and previous location of the RSR, and $\boldsymbol{x(t)_{HC}} \in R^6$ and $\boldsymbol{x(t-1)_{HC}} \in R^6$ is the current and previous location of the HC (user's hands). Also, $\boldsymbol{S} \in R^2$ is the dynamic scaling factor based on the user's input speed and is defined as

$$\boldsymbol{S} = \frac{[Max_v, \ Max_\omega]}{[\|\boldsymbol{v_{HC}}\|, \ \|\boldsymbol{\omega_{HC}}\|]} \quad (11)$$

where $\|\boldsymbol{v_{HC}}\|$ and $\|\boldsymbol{\omega_{HC}}\|$ is the norm of the linear and angular velocities of the HC (user's hands) during motion ~ $\boldsymbol{\dot{x}(t)_{HC}} \in R^6$. Also, $Max_v, \ Max_\omega$ are the desired maximum linear and angular velocities based on the user's desired input.

Additionally, the trajectories from the haptic controller ($\boldsymbol{x(t)_{HC}}$) are collected at a maximum sampling rate of 1 kHz, due to the communication protocol (USB – connection), which does not meet the required sampling rate (10 kHz) to actuate the RSR stepper motors. Given this, the incremental trajectory of the RSR ($\boldsymbol{x(t)_{RSR}}$) is up-sampled using linear interpolation. Further, we implement the Kalman filter to predict the future state of the RSR where the hidden sates of trajectories are continuous (i.e, velocity, and acceleration). In our approach, we infer the hidden states of RSR (velocity and acceleration) from the observable state, i.e., position trajectory. The Kalman filter can be modeled as

$$\boldsymbol{x(t+1)} = F\boldsymbol{x}(t) + K(t+1)\left(\boldsymbol{x(t)_{RSR}}(t+1) - HF\boldsymbol{x}(t)\right) \quad (12)$$

Where $\boldsymbol{x(t)}$ is the predicted observable state $\in R^6$, $F$ is the transition matrix, $k(t)$ is the Kalman gain, $\boldsymbol{x(t)_{RSR}}$ observations $\in R^6$, and H is the observation matrix.

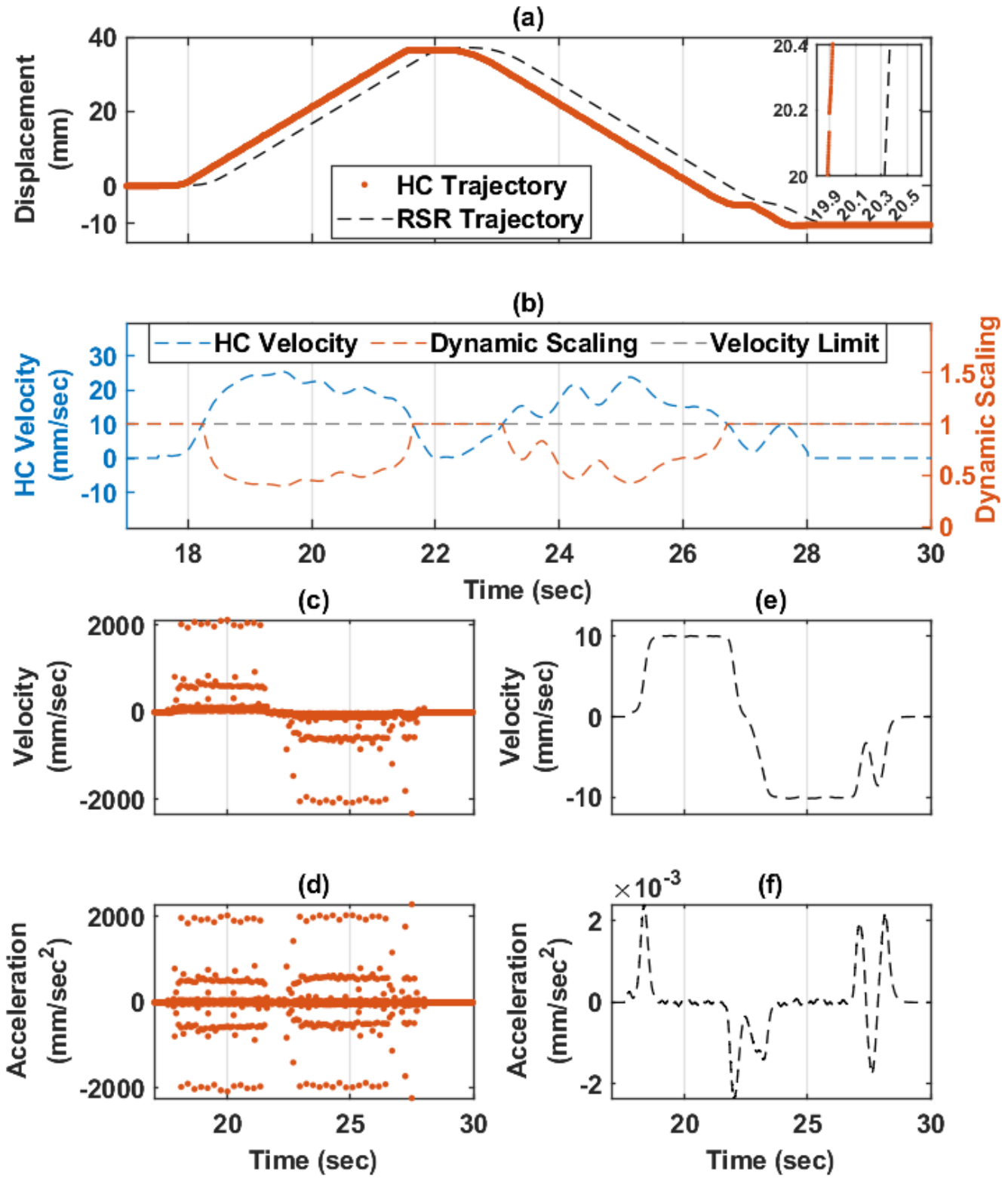

**Fig. 8.** Analysis of the motions transfer approach. The input trajectory from the HC has a breakage in position, velocity, and acceleration. **(a)** Shows the position profile from the HC and RSR desired trajectory. **(b)** The incorporated dynamic scaling factor as a function of the user's hand motion speed is presented. Once the user speed is above the desired velocity limits (10 mm/sec), the dynamic scaling factor scales the position of the trajectory accordingly. **(c & d)** shows the instantaneous velocity and acceleration profile for the original raw HC trajectory where there is a breakage in position, velocity, and acceleration. **(e & f)** shows the instantaneous velocity and acceleration profile for the RSR where the trajectory continuous in position, velocity, and acceleration . Implementation of the proposed approach results in a smooth and continuous position, velocity, and acceleration trajectory as input to the RSR. The delay between the HC and RSR trajectory < 500 ms.

**Fig. 8** illustrates the analysis of the motion integration approach proposed in this paper. The HC (hand-motion) trajectory shows a breakage in position, velocity, and acceleration (**Fig 8 a, c & d**). The high velocity and acceleration in the raw trajectory are due to the breakage in the position profile and the loss of communication between the HC and control software. We implement a moving average filter before the Kalman filter to smooth the hidden state of the trajectories (i.e, velocity, and acceleration).

After processing, we are able to generate a trajectory that is continuous for the position, velocity, and acceleration (**Fig 8 a, e & f**). Also, the trajectory hidden state (i.e., the velocity profile) is capped at 10 $\frac{mm}{sec}$ (for demonstration purposes only, in actual setting the speed is less than 2 mm/sec) despite the HC velocity (user's hand speed) moving freely at a higher speed of ~ 20 mm/sec (**Fig 8 b).**

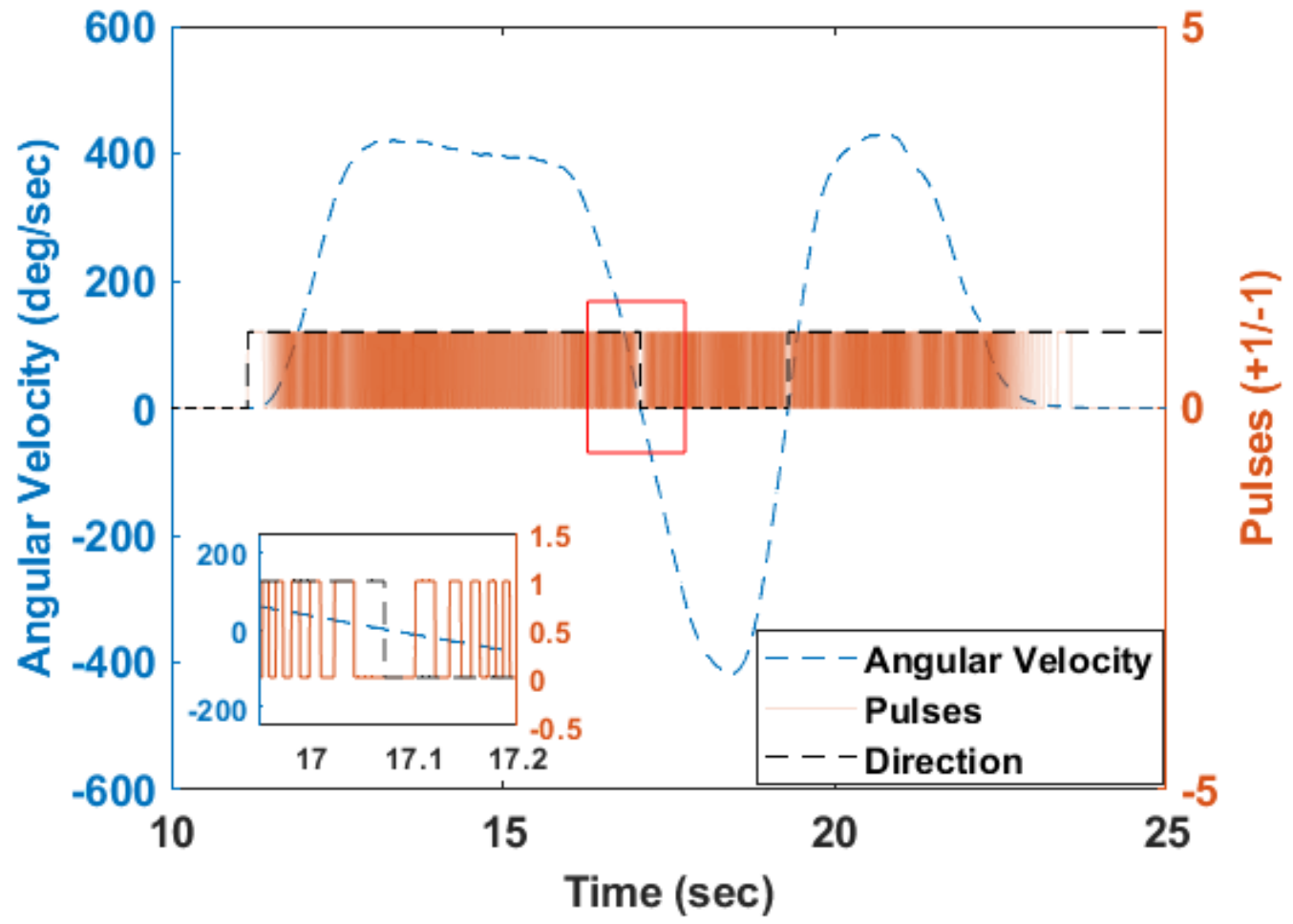

**Fig. 9.** Motor actuation of RSR during a movement in the z-direction. The frequency and period of the pulses are proportional to the angular velocity, and the direction of the rotatory shaft changes as the direction of the motion changes.

## *A. Motor Actuation*

The joint velocity of the Robossis surgical robot is a $6 \times 1$ vector denoting the three rotatory and linear actuators, respectively:

$$\dot{\boldsymbol{q}}_{RSR} = \left[\,\dot{\theta}_1\ \dot{\theta}_2\ \dot{\theta}_3\ \dot{d}_1\ \dot{d}_2\dot{d}_3\right]^T \quad (13)$$

The linear and angular velocities of the moving platform are defined to be v and ω, respectively. Thus, $\dot{x}_{RSR}$ can be written as a 6 × 1 velocity vector:

$$\dot{\boldsymbol{x}}_{\boldsymbol{RSR}} = [\boldsymbol{v_{RSR}}\ \boldsymbol{\omega_{RSR}}] \quad (14)$$

The Jacobian matrix relates $\dot{q}_{RSR}$ and $\dot{x}_{RSR}$ as follows:

$$\dot{\boldsymbol{q}}_{\boldsymbol{RSR}} = J\dot{\boldsymbol{x}}_{\boldsymbol{RSR}} \quad (15)$$

The concept of reciprocal screws is applied to derive $J$ [37]–[40]. As such, the angular velocity of each joint is determined to derive the corresponding actuator pulses. By definition, the number and frequency of the pulses for each motor are proportional to the angular velocity of the joints. We use the Euler backward numerical approximation to estimate the number of revolutions for each motor and, hence, determine the number of pulses to actuate each motor:

$$q(t)_{RSR} = q(t-1)_{RSR} + K * [t_{clock}(t) - t_{clock}(t-1)] * \dot{q}(t)_{RSR} \quad (16)$$

$$Pulses = \begin{cases} 1, \ \sin(q(t)_{RSR}) > 0 \\ 0, \ \sin(q(t)_{RSR}) < 0 \end{cases} \quad (17)$$

where k is the gain constant that is a property of each motor. **Fig. 9** shows the analysis of the motor actuation of the RSR during a motion in the z-direction. The frequency of the pulses is proportional to the angular velocity, and the direction of the rotatory shaft changes as the direction of the motion changes.

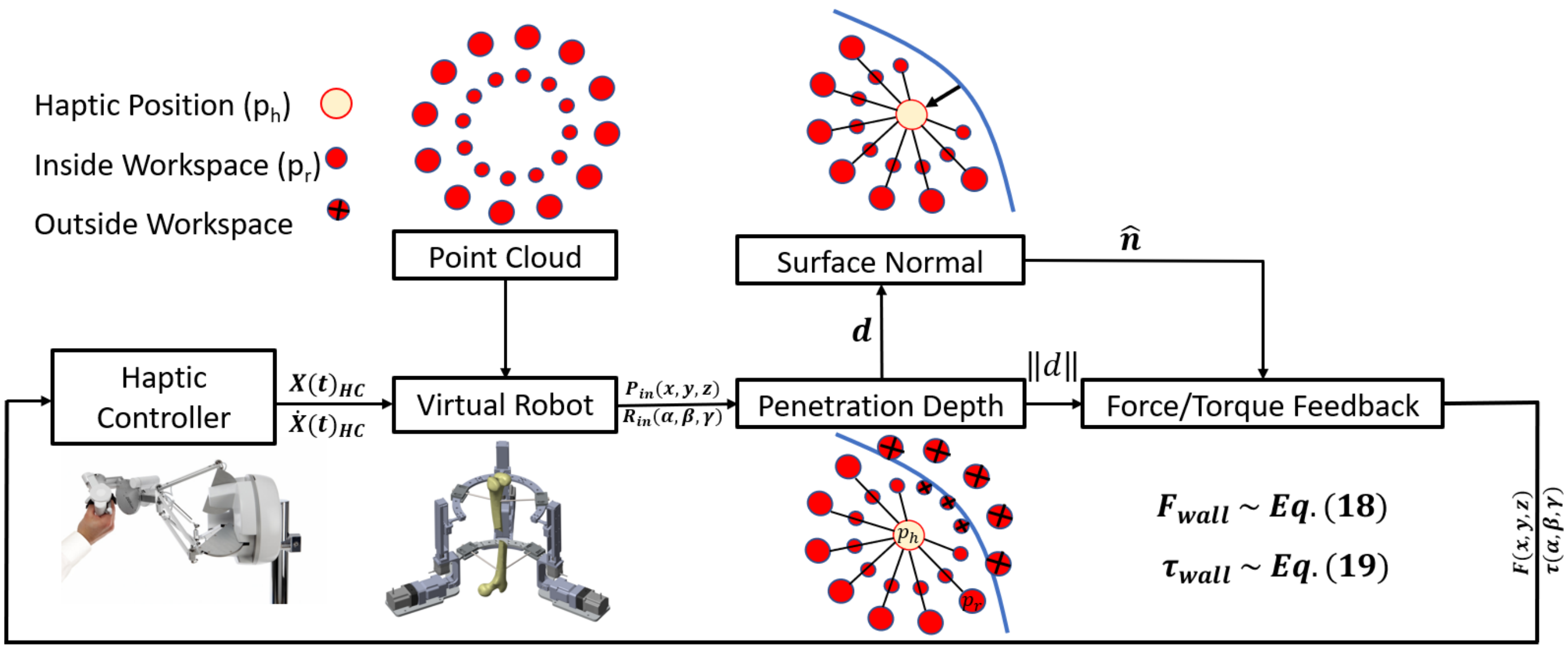

**Fig. 10.** Haptic rendering pipeline. The method used to render the motion of the HC within the RSR joint space limits. The haptic rendering pipeline restricts the movement of the HC inside a virtual workspace of the RSR which changes based on the orientation and locations of the RSR end-effector due to the nature of the robot mechanism. The end effector is encapsulated by two spherical point clouds, a translational and a rotational spherical point cloud, with radii 15 mm and 5°, respectively. When any of the points in these two spheres are outside the RSR joint space limits, the movement of the HC is restricted via haptic force feedback.

## VI. Haptic 6-DOF Joint Space Rendering

Due to the nature of the RSR parallel mechanism and coupling between the translational and rotational workspaces, the allowable workspace of the RSR changes with respect to the end effector position and orientation. Also, the leader HC is of a different mechanism than the RSR which leads to a kinematic mismatch between the mechanisms and results in damage to the systems if not accounted for in the control cycle. As such and to account for the kinematic mismatch, a haptic rendering pipeline is proposed. The haptic rendering pipeline restricts the movement of the haptic controller inside a changing virtual workspace of the RSR based on the physical constraints of the structure (**Fig. 10**).

### *A. Haptic Rendering Pipeline*

The haptic feedback imposed on the user's hand is modeled as a spring-damping system:

$$\boldsymbol{F_{wall}} = -\|d\| \cdot \boldsymbol{\hat{n}} \cdot k_f - \boldsymbol{v_{HC}} * c_f \qquad (18)$$

$$\boldsymbol{\tau_{wall}} = -\|d\| \cdot \boldsymbol{\hat{n}} \cdot k_\tau - \boldsymbol{\omega_{HC}} * c_\tau \qquad (19)$$

where $\boldsymbol{F_{wall}} \in R^3$ and $\boldsymbol{\tau_{wall}} \in R^3$ are the force and torque vectors inserted at the contact of the virtual wall, respectively. Also, $\|d\|$ is the penetration depth to the non-allowable surface which is modeled as the mean of the point clouds remaining inside the operational workspace of the RSR. $\boldsymbol{\hat{n}}$ is the unit vector; $\boldsymbol{v_{HC}}$ and $\boldsymbol{\omega_{HC}}$ are the linear and angular velocity vectors of the HC, respectively; $c_f$ and $c_\tau$ are the damping constant (15 N s/m and 0.001 Nm s/deg); and $k_f$ and $k_\tau$ are the spring constant (2000 N/m and 0.12 Nm/deg). $\|\boldsymbol{d}\| \in R^2$ is modeled as the penetration depth into the non-allowable surface of the translation and rotation workspaces and $\boldsymbol{\hat{n}} \in R^2$ is the unit vector where can be estimated as:

$$\|\boldsymbol{d}\| = \left[\left\|\frac{\sum^n \boldsymbol{P_{in}(xyz)}}{n}\right\|, \left\|\frac{\sum^n \boldsymbol{R_{in}(\alpha\beta\gamma)}}{n}\right\|\right] \qquad (20)$$

$$\boldsymbol{\hat{n}} = \frac{\boldsymbol{d}}{\|\boldsymbol{d}\|} \qquad (21)$$

where $\frac{\sum^n \boldsymbol{P_{in}(xyz)}}{n}, \frac{\sum^n \boldsymbol{R_{in}(\alpha\beta\gamma)}}{n}$ are the mean of the point clouds remaining inside the operational workspace of the RSR. The end effector of the HC is encapsulated by two spherical point clouds, a translational $\boldsymbol{P_{in}(xyz)}$ and a rotational $\boldsymbol{R_{in}(xyz)}$ spherical point cloud, with radii 15 mm and 5°, respectively.

**Algorithm 1 Haptic Rendering Pipeline**

**Haptic rendering pipeline**
**Input: Haptic Controller → $\boldsymbol{X(t)_{HC}} \in \boldsymbol{R^6}$**
***Outpoints* → []**
**For → *N # of point clouds (PC)***
**flag → Virtual Robot InvKinematics($\boldsymbol{X(t)_{HC} + PC(N)}$)**
**If flag → True (*True if (X(t)$_{HC}$ + PC(N)) is outside*)**
  ***Outpoints* → (($\boldsymbol{X(t)_{HC} + PC(N)}$)) $\in \boldsymbol{R^6}$**
**Else**
  **$\boldsymbol{P_{in}(xyz)}$ → ($\boldsymbol{X(t)_{HC,xyz} + PC(N)}$) $\in \boldsymbol{R^3}$**
  **$\boldsymbol{R_{in}(\alpha\beta\gamma)}$ → ($\boldsymbol{X(t)_{HC,\alpha\beta\gamma} + PC(N)}$) $\in \boldsymbol{R^3}$**
**End**
**End**
**If *Outpoints.Length()* > 0**
  **Determine Penetration Depth → $\boldsymbol{Eq.(20)} \in \boldsymbol{R^2}$**
  **Determine Norm of Penetration → $\boldsymbol{Eq.(21)} \in \boldsymbol{R^2}$**
  **FX, FY, FZ → $\boldsymbol{Eq.(18)} \in \boldsymbol{R^3}$ *(mean weighted)***
  **Tα, Tβ, Tγ → $\boldsymbol{Eq.(19)} \in \boldsymbol{R^3}$ *(mean weighted)***
**Else**
  **FX, FY, FZ → $\boldsymbol{-v_{HC} * c_f} \in \boldsymbol{R^3}$**
  **Tα, Tβ, Tγ → $\boldsymbol{-\omega_{HC} * c_\tau} \in \boldsymbol{R^3}$**
**End**
**Output: Force & Torque Projected on the Haptic Controller**

When any of the points in these two spheres are outside the RSR joint space limits, the movement of the HC is restricted via haptic force feedback. We implement parallel computing to determine the point clouds that are inside/outside the operational limits of the RSR workspace. As such, with this implementation, the computation cost of the haptic rendering is minimized and designed to meet the required 1 kHz haptic update loop [49]. Furthermore, we implement a mean weighted moving average on the force and torque vectors to account for the sharp changes in the surface and stability of the haptic feedback. The weights (window of 5) of the moving average is modeled as a linear function with the latest forces and torques having higher priority. Algorithm 1 shows the overall pseudocode for the haptic rendering pipeline. The input from the HC includes the mapped translation and rotation location of the end effector. Then, a check for all the point clouds determines if the points are inside or outside the operational workspace while implementing the parallel computing to optimize the run time for the computation. Furthermore, if any of the point clouds are outside the workspace, the pipeline imposes haptic feedback on the user's hands.

### *B. Haptic Feedback Analysis*

A simulation study was completed to analyze the haptic feedback magnitude and direction at the edges of the workspace. The simulation was completed by moving the end effector of the RSR into the edges of the workspace and collecting the haptic feedback responses. **Fig. 11** shows the magnitude of the forces and torques as the haptic end-effector penetrates the surfaces of the RSR's dexterous and rotational workspaces, where it was at a maximum at the edge of the boundary. The force feedback analysis shows the magnitude of the force and torque increases to a maximum (15 N and 0.3 Nm) at the edge of the boundary. Also, haptic feedback is modeled as a gradual haptic increase as the motion approaches the boundary of the workspace. The haptic feedback is normal to the penetrating vectors, with increasing forces and torques as the end effector reaches the boundary.

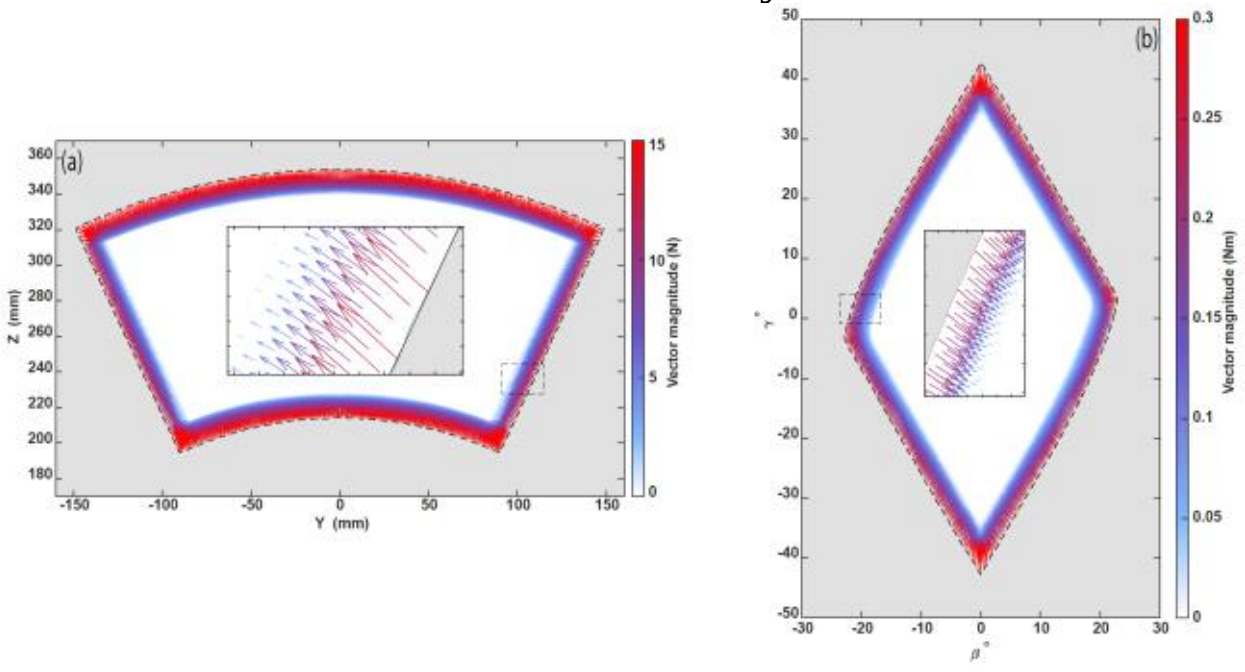

**Fig. 11.** Force feedback analysis shows the magnitude of the force and torque increases to a maximum at the edge of the boundary. Also, haptic feedback is modeled as a gradual haptic increase as the motion approaches the boundary of the workspace. The haptic feedback is normal to the penetrating vectors, with increasing forces **(a)** and torques **(b)** as the end effector reaches the boundary (dashed – black line).

Furthermore, an additional study was completed where the user was tasked to move in the workspace of the Robossis surgical robot. **Fig. 12** shows the forces and torque feedback analysis of the haptic rendering pipeline. Overall, as the user manipulates the haptic controller and reaches the limits of the linear actuator ($d_i$) $(0.284 \pm 0.07)\, m$ and spherical joints $(\pm 25^o)$, haptic feedback is projected into the user's hand to keep the user's hand within the limit of the workspace. If the user was forcing the HC to exceed the maximum limits of the workspace, 15 N and 0.3 Nm were the maximum haptic feedback would be experienced during the simulation from the HC (75% of the maximum forces and torques of the sigma.7 HC). Also, the computational cost of the haptic rendering pipeline during the simulation is analyzed. The result shows that the haptic rendering pipeline can meet the required 1 kHz haptic update loop during the simulation.

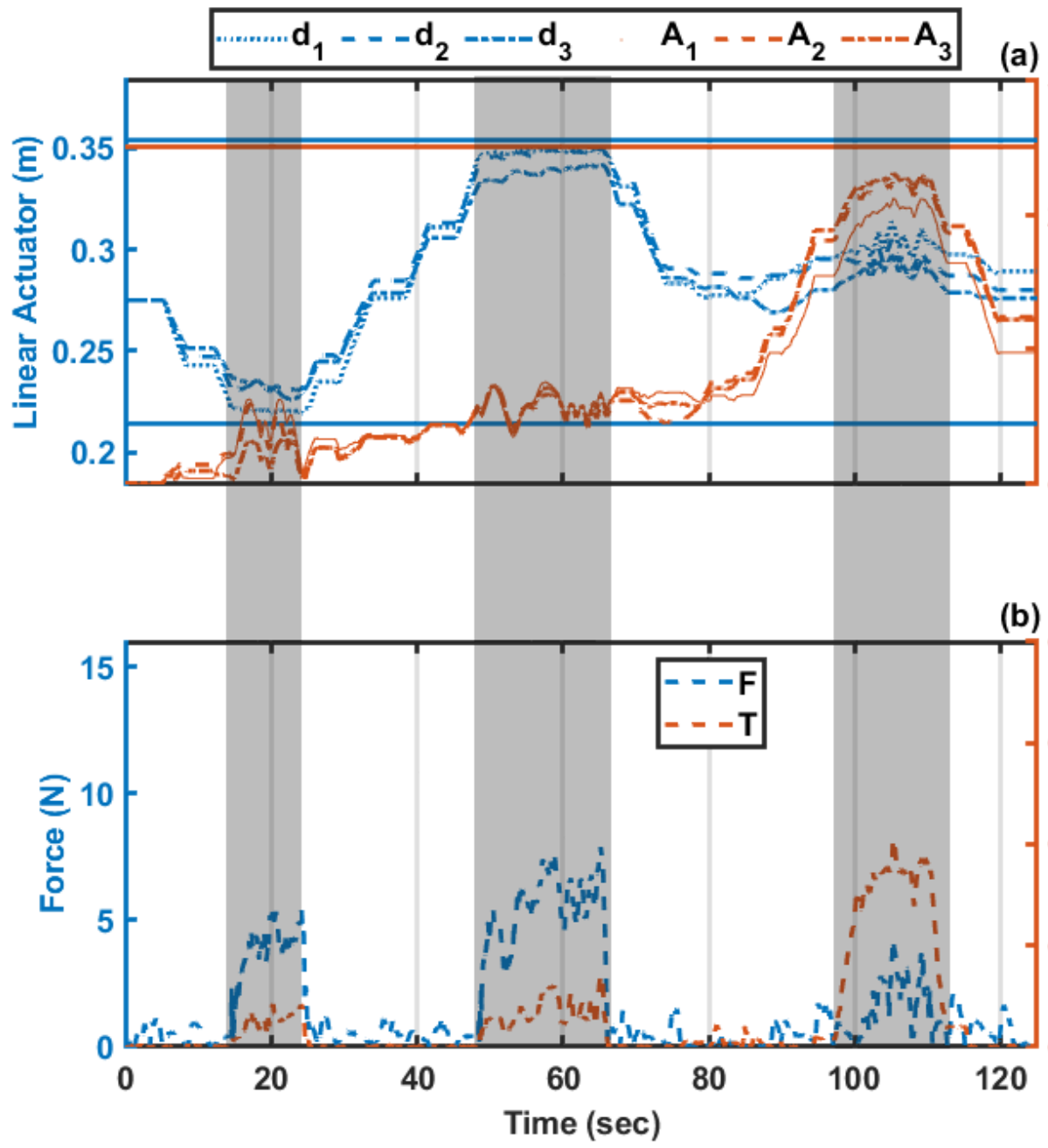

**Fig. 12. (a)** As the user manipulates the haptic controller and reaches the limits of linear actuator (0.214m-0.354m) and spherical joints $(\pm 25^o)$, **(b)** haptic feedback is projected into the user's hand to keep the user's hand within the limit of RSR joints. If the user was forcing the HC to exceed the maximum limits of the workspace, 15 N and 0.3 Nm were the maximum haptic feedback would be experienced during the simulation from the HC. When the user is not in a collision with the surfaces, a viscous force is projected onto the user hand. The user maximum speed while interacting with the haptic surfaces was 10 mm/sec and 1 deg/sec.

## VII. Experimental Testing and Results

The Robossis system was tested in a benchtop and cadaver lab as a proof of concept. **Fig. 13** shows the Robossis system, which consists of a haptic unit, a surgical unit, and the different components of the system, including sigma.7, RSR, hardware/software, and the optical tracking system (Optitrack Flex 13, residual within $\pm$ 0.08 mm and $\pm$ 0.03°, NaturalPoint, Inc. DPA OptiTrack). Optitrack's motion capture cameras and Motive software were used to track the 6-DoF motion of RSR during the experimental testing. To track the end effectors of the RSR, markers were placed in a specific location and orientation on the moving rings and three Optitrack cameras were placed above RSR to capture movements.

### *A. Real-Time Motion Analysis*

Real-time motion testing of the Robossis system is completed to determine the system's accuracy when moving to different locations in the operational workspace using the GUI and HC (video attached). During testing, the RSR was tasked to follow the motion of the user's hand and manual motion.

The theoretical trajectories from the HC (user's hand) and manual motion (GUI) were recorded and synced with the RSR actual trajectory (optical tracking software) (**Fig. 13**). **Fig. 14** shows the motion testing procedure trajectories for the translational and rotational motion while synced with the absolute error throughout the entire testing. The results in **Fig. 14** show that RSR can follow the motion of the desired trajectory while maintaining a minimal deviation. Error analysis (**Fig. 14**) shows an average translational, and rotational error of 0.32 mm and 0.07° between the RSR actual trajectory and HC-GUI desired trajectory. Maximum translational and rotational errors of 1.8 mm and 0.27° were observed in the z-direction and $\gamma$-direction, respectively.

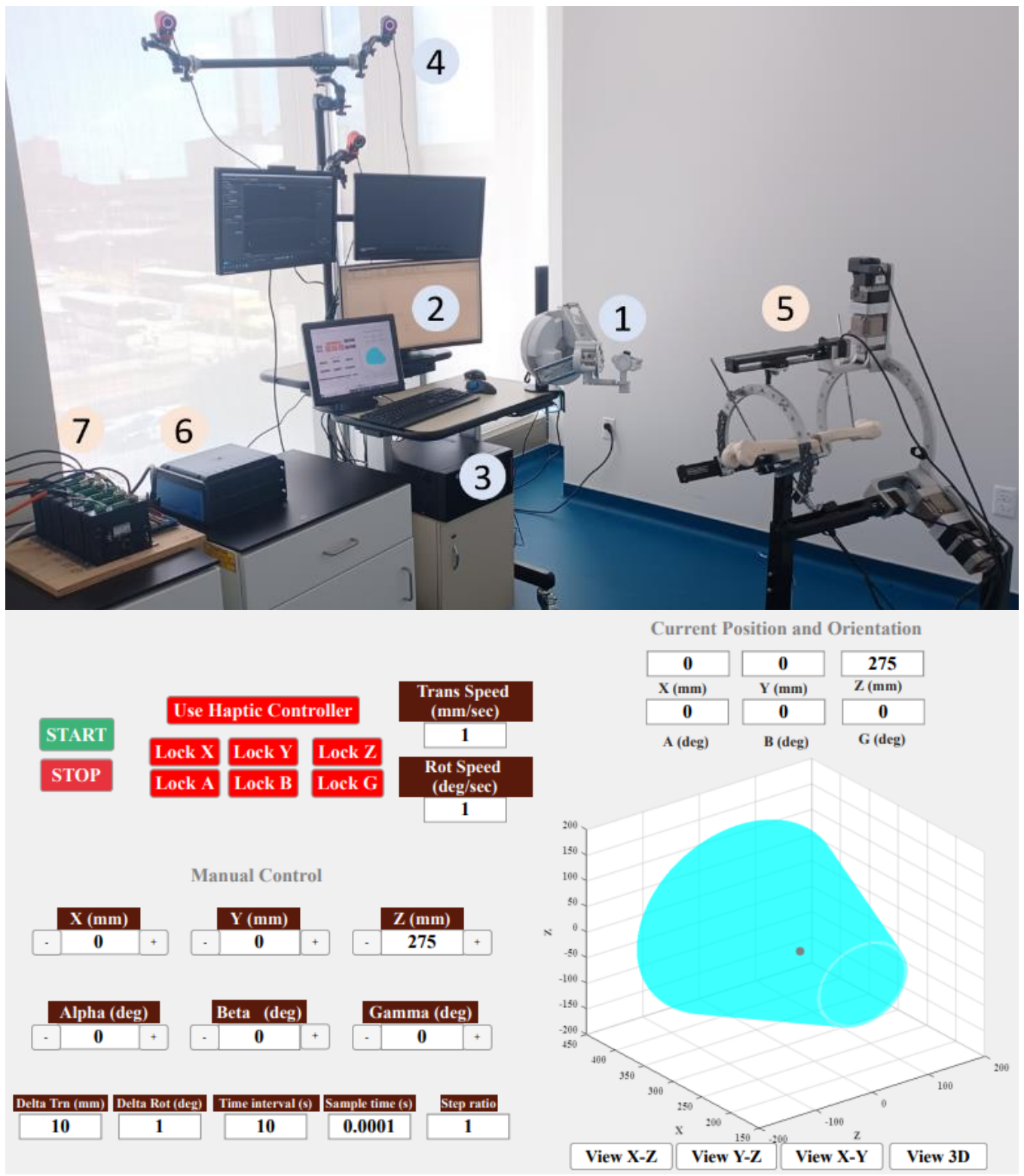

**Fig. 13.** Robossis system. The haptic unit includes (1) the haptic sigma.7, (2) screen monitors, (3) the haptic unit PC, and (4) the Optitrack optical trackers. The surgical unit includes (5) Robossis surgical robot, (6) the surgical unit PC (Speedgoat, Mathworks - Switzerland), and (7) motor drivers. The concept for the control panel graphical user interface (GUI) is presented. The user is able to manipulate the robot in all 6 DOF manually or with the haptic controller while imposing the desired time interval and speed. Current position and orientation can be viewed numerically or visually.

*B. Cadaver Lab Testing*

In a cadaveric lab, the Robossis system was used to assist surgeons during a mock femur fracture surgery. **Fig. 15 (a)** shows the cadaver lab setup, which includes the HC, RSR, and the cadaver patient attached to the RSR. During the mock surgery, the patient's femur was cut using a reciprocating saw to represent a midshaft femur fracture. The distal fragment was attached to the RSR moving ring, and the proximal fragment was clamped down to eliminate movement.

The orthopedic surgeons utilized the Robossis system to manipulate the RSR. Throughout the surgical procedure, the surgeons completed varying maneuvers to assess the Robossis system load-holding capacity (muscle traction forces) and usability in the clinical setting. The results showed that Robossis was able to assist the surgeon in performing femur fracture surgery and aligning the broken femur fragments (**Fig 15 (b) & (c)**). In addition, the RSR was able to counteract the actual physiological muscle traction forces while manipulating the distal bone fragments during the surgery. In [44], we present clear description of the cadaver lab experimental testing as assisted by the novel imaging software. Lastly, the surgeons were able to conclude that the Robossis system is able to provide an intuitive solution for surgeons' to perform femur fracture surgery.

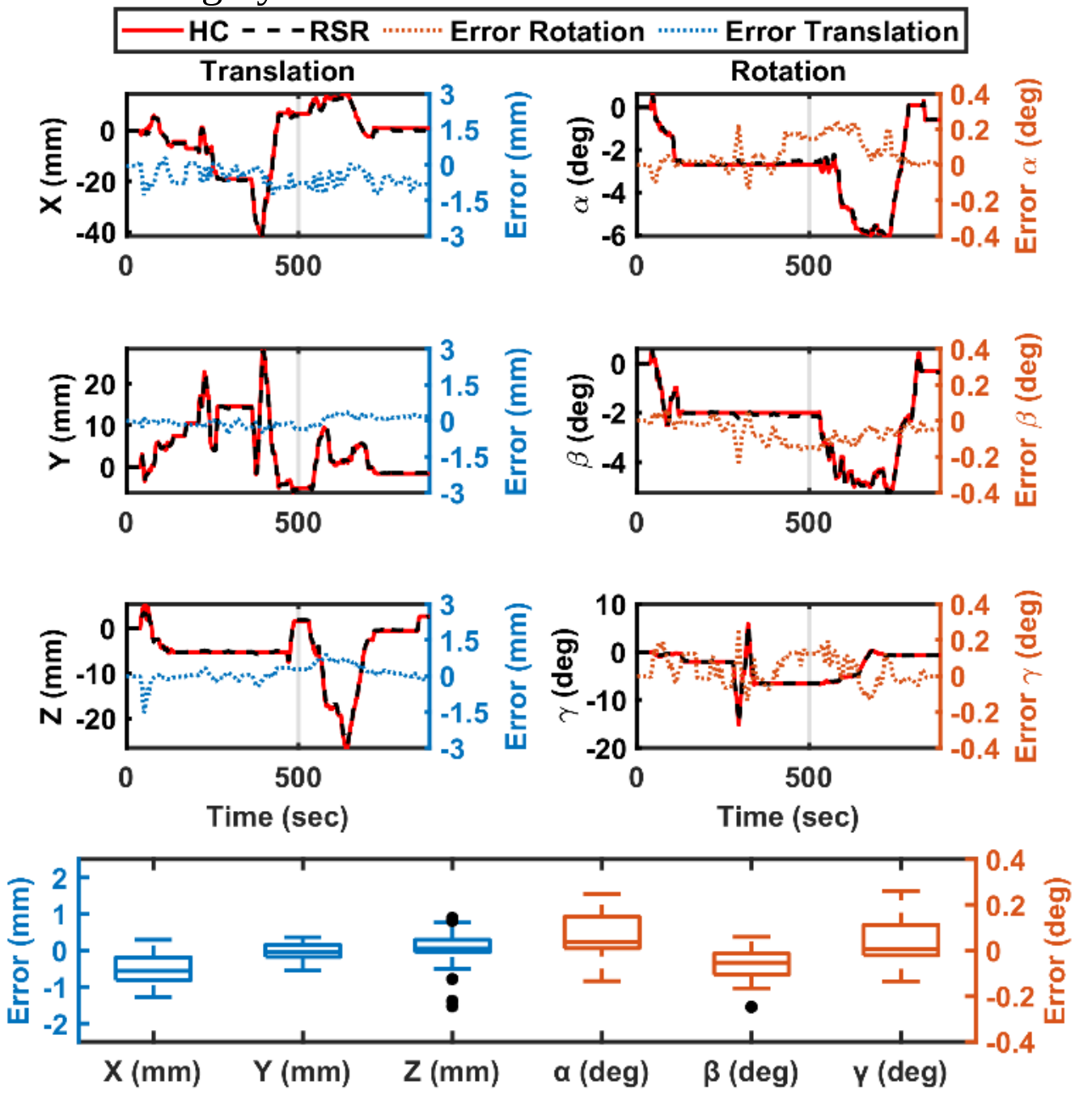

**Fig. 14.** Trajectory comparison between the RSR actual trajectory (Optitrack) and theoretical trajectory from the HC and GUI. The motion testing procedure trajectories for the translational and rotational motion were synced with the absolute error throughout the entire testing. The box and whisker plot shows an average translational, and rotational error of 0.32 mm and 0.07° between the RSR actual trajectory and HC-GUI desired trajectory.

## VIII. Discussion

The outcome of this study demonstrates the feasibility of the proposed Robossis system in assisting surgeons' during long-bone femur fracture surgery. With this ability, the Robossis system provides a significant advantage over the current conventional surgical techniques and prior attempts at robotic developments [23], [24]. Considering the Robossis system in the application of femur fracture, the surgical robot needs to be highly maneuverable, insert the required forces, and have highly accurate manipulation. In contrast, manual manipulation and a limited workspace will prevent the patient from being positioned at the surgeon's discretion, leading to higher risks of patient injury and a decrease in surgical performance. Also, the integration of the haptic system further demonstrates the advantages over previous attempts as it provides an intuitive system for the surgeon's.

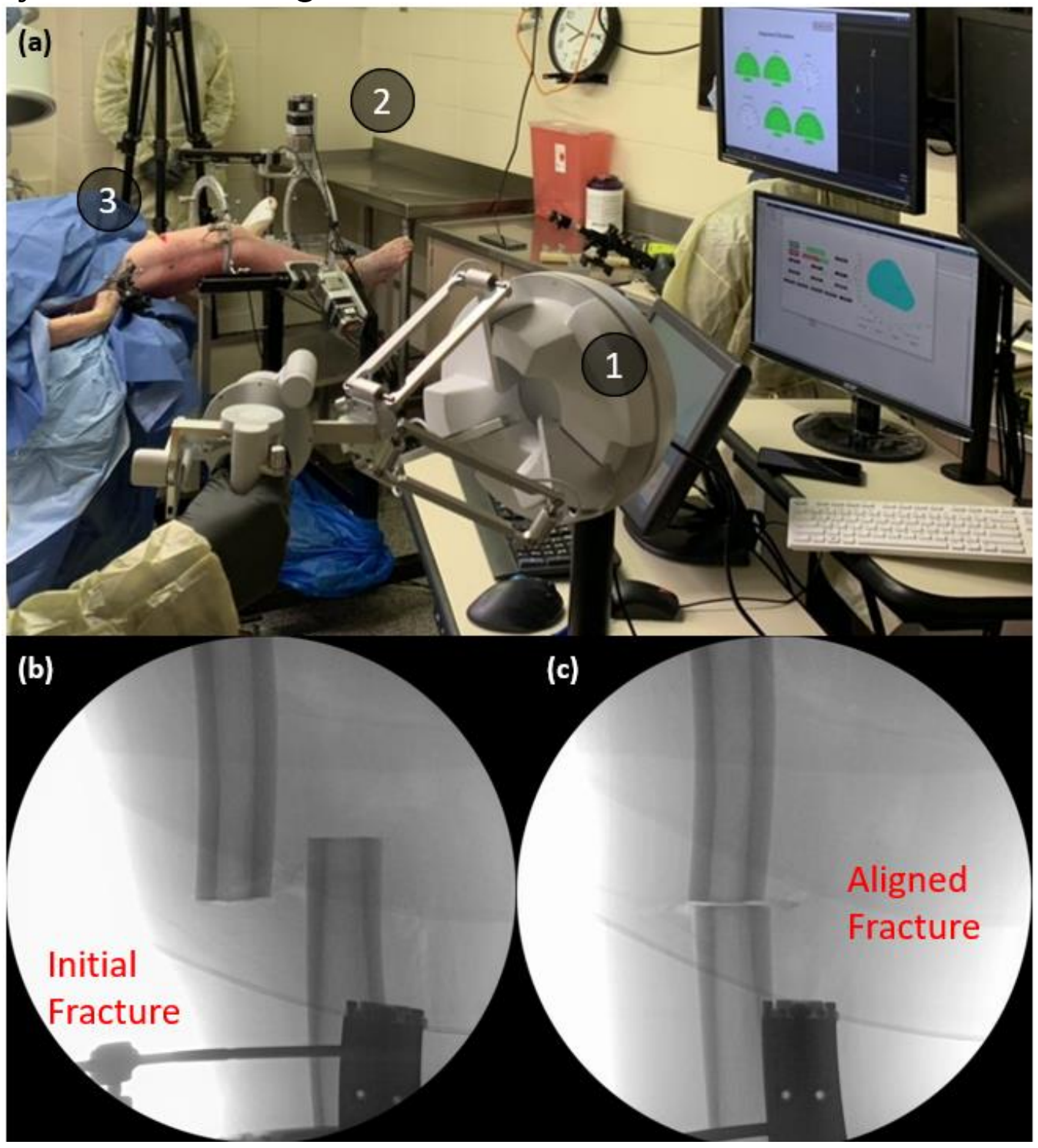

**Fig. 15. (a)** Robossis system was used to assist the orthopedic surgeons during a cadaveric femur fracture surgery where (1) the haptic sigma.7, (2) the RSR attached to the cadaver patient (3). **(b)** shows the femur pre-fractured location of the bone. **(c)** shows the aligned fracture using the Robossis system.

Furthermore, the theoretical simulation validates the inverse kinematic model of the RSR to manipulate the end effector (moving ring) given the desired translational and rotational trajectories. Despite the minor errors in the simulation (~ 0.2 mm and ~ 0.1 deg), which are attributed to the Gazebo simulation engine in keeping the linear actuators' length and active joints' angles into desired, we believe this simulation reinforces the working principles of the mechanism and the relationship between user desired motion and input trajectory.

Also, the kinematic simulation, trajectory analysis, and benchtop experimental testing demonstrate the accuracy of the kinematic interface between the HC and RSR. The results reinforce the reliability of the control software and its inverse kinematics model to deliver the proper joint velocities to move precisely in the translational and rotational directions, which is required for surgical performance during motion replication of the RSR [50]. Also, the dynamic scaling of the trajectory restricted the speed of the input trajectory to the desired maximum velocity. Given this, the velocity of the RSR is restricted to eliminate any possible speed-induced tissue damage during the surgery or further complication while also reducing the impact of surgical tremors [51]. The error during the actual experimental testing is attributed due to the motion capture system, tracking the markers within mm accuracy, and joint clearances during the machining of the joints. The error is anticipated to be reduced after design and manufacturing improvements are implemented. Also, including more motion-capturing cameras would minimize the source error in tracking the markers for future testing.

Furthermore, the haptic rendering pipeline is implemented to keep smooth and continuous motion within the RSR's operational workspace limits. This ensures the safety of the RSR's joints and eliminates any chances of failure during the surgery. At contact with the changing virtual surfaces, the direction of the force feedback is normal to the surface, with a magnitude increasing as the penetration into the surface increases. As such, the motion at the boundary is smooth and allows the surgeon to reach all possible points in the dexterous and rotational workspaces. Also, the computational overload of the haptic rendering pipeline is minimized by implementing the parallel computing algorithm. This increases the computational integrity of the system to manage other components, such as motion transfer and motor actuation. We believe the haptic rendering pipeline can be generalized to the varying surgical robotic mechanism and eliminate the challenges of integrating a kinematically dissimilar leader-follower system.

The highlighted challenge of the proposed haptic rendering pipeline is the speed at which the virtual surfaces are changing which is proportional to the user speed while interacting with the HC. This challenge is because the magnitude of the haptic feedback is proportional to the speed at which the virtual surfaces are changing. Although, the Robossis system is designed to move at a low speed (2 mm/sec and 0.2 deg/sec) which eliminates the highlighted challenge.

In learning the current abilities and limitations of the proposed Robossis system, we seek to implement improvements in future work. We will improve on the current unilateral architecture to integrate haptic feedback from the RSR. In this, we will design a bilateral architecture where the surgeon will have a better feeling of RSR-patient interaction during the surgical procedure. Furthermore, the current system requires experience and training to properly improve the standard of care for long-bone fractures. The next step will be implementing a VR simulation to allow surgeons to feasibly train and master the system.

## IX. Conclusion

To successfully restore the length, alignment, and rotation of the fractured femur, the femur fragments must be manipulated and returned to their correct anatomical position. All of this must be done while the surgeon is exerting large traction forces and torques (517 N and 74 N·m). In this study, we have been able to present the development of the Robossis system for femur fracture surgery that includes the design of a unilateral haptic architecture that addresses (1) the kinematic mismatch and (2) real-time motion transfer between the HC and RSR.

The feasibility of the Robossis system was experimentally evaluated through a benchtop and cadaveric experiment. Through experimental testing and clinician feedback, we demonstrated that the system has the potential for clinical use to improve the quality of fracture reduction and alignment.